\documentclass[letterpaper, 10 pt, conference]{ieeeconf}  % Comment this line out if you need a4paper
\IEEEoverridecommandlockouts

\usepackage{amsmath}

\usepackage{cleveref}
\usepackage{cite}

\usepackage{pgf}
\usepackage{cite}
\usepackage{wrapfig}
\usepackage{amsmath,amssymb,amsfonts}
\usepackage{algorithmic}
\usepackage{graphicx}
\usepackage{textcomp}
\usepackage{xcolor}
\usepackage{booktabs}

\usepackage{graphicx}
\usepackage{diagbox}
\usepackage{tabularx}

\usepackage{cite}
\usepackage{amsmath,amssymb,amsfonts}
\usepackage{algorithmic}
\usepackage{graphicx}
\usepackage{textcomp}
\usepackage{xcolor}

\usepackage{gensymb} % For degree symbol

\usepackage{multirow}
\usepackage{dcolumn}
\usepackage{diagbox}

\usepackage{multirow}
\usepackage{dcolumn}
\usepackage{makecell}
\usepackage{graphicx}
\usepackage{subcaption}
\usepackage{color,soul} %\sout
\definecolor{red}{rgb}{1.00,0.00,0.00}
\definecolor{blue}{rgb}{0.00,0.00,1.00}
\definecolor{green}{rgb}{0.30, 0.50,0.00}

\usepackage{multirow}

\def\BibTeX{{\rm B\kern-.05em{\sc i\kern-.025em b}\kern-.08em
    T\kern-.1667em\lower.7ex\hbox{E}\kern-.125emX}}

\begin{document}

\title{HMT-Grasp: A Hybrid Mamba-Transformer Approach \\ for Robot Grasping in Cluttered Environments}

\author{Songsong Xiong$^{1}$, Hamidreza Kasaei$^{1}$% <-this % stops a space
\thanks{$^{1}$Department of Artificial Intelligence,
        University of Groningen, Groningen, The Netherlands\newline 
        {\tt\small \{s.xiong, hamidreza.kasaei\}@rug.nl}}}

\maketitle
\thispagestyle{empty}
\pagestyle{empty}

%%%%%%%%%%%%%%%%%%%%%%%%%%%%%%%%%%%%%%%%%%%%%%%%%%%%%%%%%%%%%%%%%%%%%%%%%%%%%%%%
\begin{abstract}
Robot grasping, whether handling isolated objects, cluttered items, or stacked objects, plays a critical role in industrial and service applications. However, current visual grasp detection methods based on Convolutional Neural Networks (CNNs) and Vision Transformers (ViTs) often struggle to adapt to diverse scenarios, as they tend to emphasize either local or global features exclusively, neglecting complementary cues. In this paper, we propose a novel hybrid Mamba-Transformer approach to address these challenges. Our method improves robotic visual grasping by effectively capturing both global and local information through the integration of Vision Mamba and parallel convolutional-transformer blocks. This hybrid architecture significantly improves adaptability, precision, and flexibility across various robotic tasks. To ensure a fair evaluation, we conducted extensive experiments on the Cornell, Jacquard, and OCID-Grasp datasets, ranging from simple to complex scenarios. Additionally, we performed both simulated and real-world robotic experiments. The results demonstrate that our method not only surpasses state-of-the-art techniques on standard grasping datasets but also delivers strong performance in both simulation and real-world robot applications.

\end{abstract}

\section{Introduction}
\label{sec:intro}

Robotic grasping is a fundamental capability for autonomous manipulation in industrial automation, service robotics, and domestic applications~\cite{fu2024light, liu2021robotic}. Robots operate in real-world situations where objects vary in shape, size, and material properties and are randomly arranged and partially occluded. In such challenging conditions, robust and adaptive grasping strategies are needed to identify feasible grasp configurations despite significant visual and physical clutter~\cite{zhao2022robot}.

Conventional approaches to robotic grasping relied on analytical methods and hand-crafted features, which performed adequately under controlled conditions but failed to generalize to complex, unstructured environments. With the advent of deep learning, convolutional neural networks (CNNs) have become the predominant paradigm for extracting discriminative local features that are essential for grasp detection~\cite{jiang2011efficient, lenz2015deep, redmon2015real, morrison2020learning, kumra2020antipodal}. However, in cluttered scenes, local features alone are often insufficient because grasping decisions depend critically on global contextual cues—such as spatial relationships among objects and prevalent occlusion patterns—which traditional CNNs may not adequately capture. Recently, transformer architectures have underscored their ability to model long-range dependencies and provide a more comprehensive contextual representation~\cite{wang2022transformer,dong2022robotic}. Nonetheless, these methods may inadvertently diminish the local spatial details—such as edges, corners, and shapes—that are crucial for accurate grasp detection~\cite{raghu2021vision}. In contrast, recent advances in Mamba architectures extract both local and global features in a more balanced manner than ViTs and CNNs~\cite{xu2024survey,ruan2402vm}, which tend to favor one over the other, thereby mitigating their limitations.

To improve robot grasp detection performance in multiple scenarios, we propose a novel Mamba-Transformer architecture that leverages a state space model (SSM) to fuse and refine features extracted by parallel CNN and Transformer modules. By effectively integrating local detail with global contextual cues, our approach facilitates robust grasping detection across various scenes. Experimental results show that our method outperforms state-of-the-art techniques on multiple grasping datasets and delivers superior performance in both simulation and real-world robotic applications.
The main contributions of this paper are summarized as follows:

\begin{enumerate}
\item We propose a hybrid Mamba-Transformer architecture that integrates parallel CNN and Transformer streams, leveraging their respective strengths to enable the simultaneous extraction of fine-grained local features and global contextual cues for grasp detection.
\item We then present the Mamba encoder to further fuse local and global features, which refines fine local details while effectively integrating broader contextual information.
\item We conduct comprehensive evaluations and rigorous ablation studies across multiple grasping datasets, achieving excellent performance. Additionally, we carry out extensive simulation and real robot experiments. The experimental results clearly demonstrate that our approach exhibits robust performance.
\end{enumerate}

\begin{figure*}[!t]
\centering
\includegraphics[width=15cm,height=6cm]{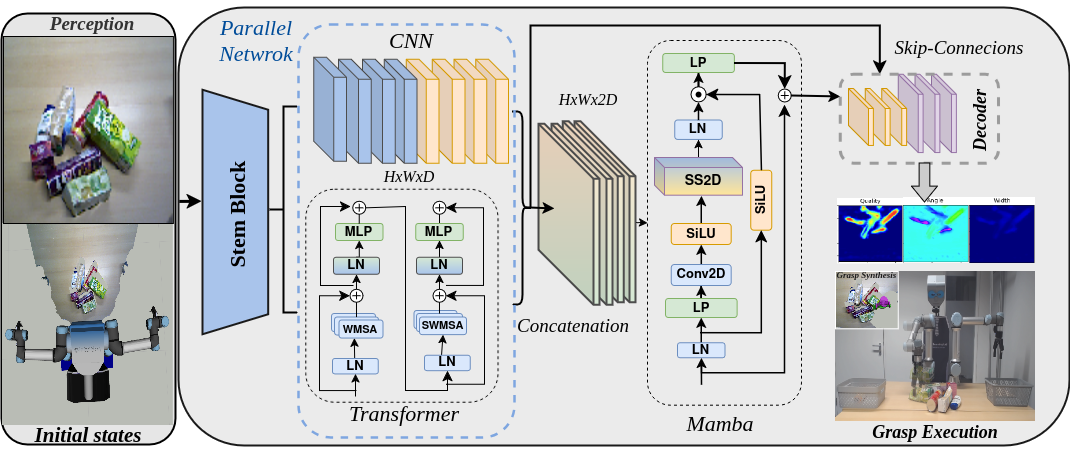}
\caption{Overview of the hybrid mamba-transformer architecture for robotic grasping in cluttered environments: The system comprises a grasp encoder, which integrates parallel CNN and transformer networks, followed by mamba blocks for enhanced feature fusion. A grasp decoder with skip connections further refines local and global feature extraction. The decoder upsamples to predict grasp quality, angle, and width, forming the grasp synthesis. Finally, based on the synthesized grasp prediction, the robot executes the grasp. }
\label{fig:architecture}
\vspace{-0.5mm}
\end{figure*}

\section{Related Work}
This section mainly reviews recent advancements in robotic grasp detection. Current research on robot grasp detection is primarily categorized into two approaches: CNN-based methods and ViT-based methods. Additionally, we provide a brief overview of recent progress with Vision Mamba in various domains.

\subsection{Grasp Detection}

\noindent\textbf {CNN-Based Methods}: 
Convolutional Neural Networks (CNNs) have been widely applied in robotic visual grasping tasks~\cite{morrison2020learning}. Early work by Lenz et al. introduced a two-step cascaded system that employed two deep neural networks to predict the grasp pose of objects. In this approach, numerous candidate grasps are first generated and then re-evaluated in a second step, which significantly slows down the detection process~\cite{wang2022transformer}. Subsequent research has leveraged CNNs to enhance object grasping performance by predicting grasp rectangles. Redmon et al. proposed a single-stage regression model that enabled real-time grasp detection~\cite{redmon2015real}. To further improve accuracy, a ResNet-50-based grasp detector was introduced, demonstrating superior performance on the Cornell RGB-D dataset~\cite{kumra2017robotic}. Datasets such as Jacquard~\cite{depierre2018jacquard} and Multi-Object~\cite{chu2018real} have contributed to enhancing the accuracy and generalization of CNN models in grasp detection. In addition, the Generative Grasping CNN (GG-CNN), designed to work with depth images, improved pixel-wise grasp quality in real-time scenarios. GR-ConvNet~\cite{kumra2020antipodal}, proposed by Kumra et al., used n-channel inputs to detect antipodal grasps, achieving high accuracy on the Cornell and Jacquard datasets. Despite the effectiveness of CNN-based methods in grasp detection, a significant drawback is their tendency to lose global context due to the inherent structure of stacked convolutional layers~\cite{wang2022transformer}.

\noindent \textbf{ViT-Based Methods}: 
Motivated by the success of Vision Transformers (ViTs) in object detection and image classification, researchers are increasingly exploring their potential for predicting object grasp poses in robotic tasks~\cite{wang2022transformer,dong2022robotic,han2021learning,yenicesu2023fvit}. Wang et al.~\cite{wang2022transformer} were among the first to propose the use of Swin-Transformers for visual grasp detection, employing them as both encoders and decoders with skip connections to generate pixel-level grasp representations. Building on this, Dong et al.~\cite{dong2022robotic} also utilized Swin-Transformers for encoding, coupled with fully convolutional decoders to enhance grasp detection performance. In a more recent effort, Yenicesu et al.~\cite{yenicesu2023fvit} introduced LITV2, a lightweight ViT model, to achieve real-time grasp detection. Compared to CNN-based methods, ViT-based approaches have demonstrated improved grasp detection accuracy. However, as the depth of transformer blocks increases, local-level information tends to gradually diminish~\cite{raghu2021vision}.

\subsection{Visual Mamba}

Recently, Mamba, a novel State Space Model (SSM) initially developed for natural language processing (NLP)~\cite{gu2312mamba}, has shown strong potential for efficiently modeling long sequences. This success has sparked increasing interest in applying SSMs to visual tasks. In particular, Mamba has been successfully implemented in various computer vision applications. For example, ViM~\cite{zhu2024visionmambaefficientvisual}, a Mamba-based architecture, operates on image patch sequences using a 1D causal scanning mechanism. However, the 1D scanning approach faces challenges in fully capturing global representations. To address this limitation, VMamba~\cite{liu2024vmambavisualstatespace} introduces 2D-Selective-Scan, transforming images into sequences of patches for improved global feature extraction. 

Inspired by these advancements, SegMamba~\cite{xing2024segmambalongrangesequentialmodeling} and VM-UNet~\cite{ruan2024vmunetvisionmambaunet} have successfully integrated visual state space models into encoders for medical image segmentation, contributing to notable performance improvements in this domain. Despite these developments, no prior work has explored the application of SSMs in robotic grasp detection. Given the strengths of mamba in capturing both global information and local details, alongside the complementary capabilities of transformers and CNNs, we designed a Mamba-Transformer approach to predict robotic grasp poses. Our method demonstrates superior grasp success rates compared to state-of-the-art transformer- and CNN-based approaches.

\section{Method}

In this section, we present a hybrid Mamba-Transformer architecture for robotic grasp detection. The overall structure of the proposed model is illustrated in Fig.~\ref{fig:architecture}.

\subsection{Grasp Encoder}
We first apply a Stem Block comprising two convolutional layers to downsample the input tensor. The resulting feature map is then fed into a parallel network architecture that integrates CNNs and Transformers, as shown in Fig.~\ref{fig:architecture}.

\subsubsection{Parallel Network}
To capture spatial relationships among objects, we employ the Swin Transformer block~\cite{liu2021swin}, which leverages window-based multi-head self-attention (WMSA) and shifted window-based multi-head self-attention (SWMSA). Although window-based attention extracted localized features, shifting windows promote cross-window interactions, thereby enhancing global modeling capability~\cite{raghu2021vision}. However, Transformers alone tend to suppress fine-grained local features at increasing depths~\cite{raghu2021vision}. To address this limitation, we incorporate convolutional neural networks (CNNs) within the Parallel Network. More specifically, we adopt a dual convolutional architecture~\cite{ronneberger2015u} to capture localized details such as edges and shapes. This complementary design capitalizes on CNNs’ inherent translation invariance and inductive biases, leading to more robust representations for robotic grasp detection.

\subsubsection{Mamba Blocks}
The Mamba Block employs a recursive state space model (SSM) that supports continuous information flow across the spatial domain. Two core mechanisms drive this design: \textit{implicit attention}, which propagates long-range dependencies, and \textit{dynamic filtering}, which adaptively fuses local and global features.
When input features combine CNN-derived local details and transformer-based global context, Mamba’s dynamic filtering effectively balances these distinct feature types. Consequently, Vision Mamba achieves an optimal trade-off between preserving local detail and modeling broader structures. This capability enhances the overall feature representation capacity within the Parallel Network.

Following the parallel feature extraction, we concatenate the outputs into \({x}_{m}^{l-1}\) of dimension \((B, 2D, W, H)\). Inspired by the VM-UNet architecture~\cite{ruan2024vmunetvisionmambaunet}, we feed this concatenated feature map into the \textit{mamba block}, which comprises \textit{LayerNorm} (\textit{LN}), \textit{Linear Projection} (\textit{LP}), \textit{Conv2d}, \textit{SiLU}, and the 2D-Selective-Scan (\textit{SS2D}) mechanism:
\begin{equation}
\begin{aligned}
{{x}_{m}^{l,0}}  &= \mathrm{\textit{LN}}\bigl({x}_{m}^{l-1}\bigr), \\
{x}_{m}^{l,1} &= \mathrm{\textit{LN}}\Bigl(\mathrm{\textit{SS2D}}\bigl(\mathrm{\textit{SiLU}}\bigl(\mathrm{\textit{Conv2d}}\bigl(\mathrm{\textit{LP}}({x}_{m}^{l,0})\bigr)\bigr)\bigr)\Bigr), \\
{{x}_{m}^{l}} &= \mathrm{\textit{LP}}\Bigl(\mathrm{\textit{SiLU}}\bigl({x}_{m}^{l,0}\bigr) \odot {x}_{m}^{l,1}\Bigr) +  {x}_{m}^{l-1}.
\end{aligned}
\end{equation}

where \(\odot\) indicates element-wise multiplication. The residual connection ensures stable gradient flow, while SS2D promotes richer spatial feature extraction.

\subsection{Grasp Decoder}
In the decoder phase, skip connections establish feature reuse by combining encoder outputs with corresponding decoder stages, as shown in Fig.~\ref{fig:architecture}. We apply average pooling to the encoder feature maps, then upsample them by a factor of two. A subsequent \textit{DoubleConv} operation reduces channel dimensionality:

\begin{equation}
\begin{aligned}
x_{de}^{up+1} = \mathrm{\textit{DoubleConv}}\bigl(\mathrm{\textit{UpSample}}(x_{de}^{up})\bigr).
\end{aligned}
\end{equation}

Through iterative refinement, the decoder progressively restores spatial resolution. Once feature maps reach \((\mathrm{B}, 48, 112, 112)\), they are upsampled again to yield pixel-level heatmaps matching the original image resolution of \(224\times 224\). These heatmaps include a grasp confidence map \(Q\), a gripper angle map \(\Theta\), and a gripper width map \(W\).

\subsection{Grasp Synthesis}

A robotic antipodal grasp can be defined by the tuple, $g_{i} = \bigl(\langle u, v\rangle, \phi, w, q_{i}\bigr)$
where \(\langle u,v\rangle\) denotes the grasp center in pixel coordinates, \(\phi \in \bigl[-\tfrac{\pi}{2}, \tfrac{\pi}{2}\bigr]\) is the in-plane rotation, \(w\) is the gripper width, and \(q_{i}\in[0,1]\) indicates the predicted grasp success confidence. Once the pixel \(\langle u, v\rangle\) maximizing \(q_{i}\) is found, its location is converted to Cartesian coordinates via known camera intrinsics and a calibrated transformation from the camera to the robot reference frame.

Next, the depth information at \(\langle u, v\rangle\) is extracted to determine the object’s position along the optical axis. To ensure robust execution of the grasp, the system may also consider a small neighborhood around the selected pixel to accommodate variations in sensor noise or object geometry. Finally, the manipulator is instructed to close the gripper at the computed position and orientation, thereby completing the grasp action.

\section{Experiments}

In this section, we present a comprehensive evaluation of our method through experiments on the Cornellt~\cite{kumra2017robotic}, Jacquard~\cite{depierre2018jacquard}, and OCID-Grasp~\cite{ainetter2021end} datasets. We evaluate the impact of skip connections, perform an ablation study on multiple components, and compare our approach against state-of-the-art methods. Additionally, we validate our method in both simulation and real-world robotic experiments. The grasp network has been implemented using PyTorch 2.0.1 with cuDNN 8.5.0.96 and CUDA 11.8. The model is trained end-to-end on an Nvidia RTX 4060 Ti GPU with 16GB of memory.

\subsection{Dataset and  Experiment setup }

This section outlines the datasets used in the evaluation and provides relevant training details.

\subsubsection{Cornell Dataset and Training}

The Cornell dataset is one of the most widely used benchmarks for robotic grasping. It consists of 885 images of 250 real, graspable objects with a total of 8019 annotated grasps. The dataset contains RGB-D images and can be split using two different methods: image-wise and object-wise. Given the relatively small size of the Cornell dataset, we employ 5-fold cross-validation to train and evaluate our model. To ensure robust performance, data augmentation is applied during training. The model is optimized using the Adam optimizer, with a learning rate initialized at 0.0002 and a batch size of 8. Training is carried out over 100 epochs to obtain the final model weights.

\subsubsection{Jacquard Dataset and Training}

The Jacquard grasp dataset is a large-scale dataset designed for robotic grasping tasks, comprising over 54,000 synthetic RGB-D images of more than 11,000 distinct objects. It includes approximately 1.1 million labeled grasps, generated using a physics-based simulation environment, providing diverse and comprehensive data for training grasping models. We employ image-wise splitting to train and test our network, where 90\% of the data is used for training, and the remaining 10\% is reserved for testing. While we utilize the Adam optimizer, similar to the Cornell dataset training, we adjust the learning rate to 0.001 and set the batch size to 8. The network is trained over 200 epochs to obtain the final model weights.

\subsubsection{OCID Dataset and Training}

The OCID-Grasp dataset is a large-scale, real-world dataset created for robotic grasping in cluttered environments. It contains 1763 selected images with over 75,000 hand-annotated grasp candidates, classified into 31 object categories. The dataset is designed to capture challenging grasping scenarios involving a variety of objects in cluttered settings. We utilize 5-fold cross-validation, as in the Cornell dataset training, to train the networks and report average performance across the folds. The Adam optimizer is employed with a learning rate of 0.001 and a batch size of 8. The network is trained for 80 epochs to achieve the final model weights.

\subsection{ Evaluation Metrics}

Similar to previous studies~\cite{morrison2020learning, kumra2020antipodal, wang2022transformer}, this work evaluates grasp detection using the following criteria:

\begin{itemize}
    \item Angle Difference: The orientation difference between the predicted grasp and the ground-truth is less than 30$\degree$.
    \item Jaccard Index: The Jaccard index between the predicted grasp and the ground-truth is greater than 25\%.
\end{itemize}

\subsection{Performance Comparison with State-of-the-Art Methods}
\subsubsection{Grasping Performance on the Cornell Dataset}

To demonstrate the effectiveness of our approach, we compared it against state-of-the-art methods under consistent training and testing settings across the Cornell, Jacquard, and OCID-Grasp datasets.

\begin{figure}[!t]
\vspace{1.5mm}
\centerline{\includegraphics[width=\linewidth]{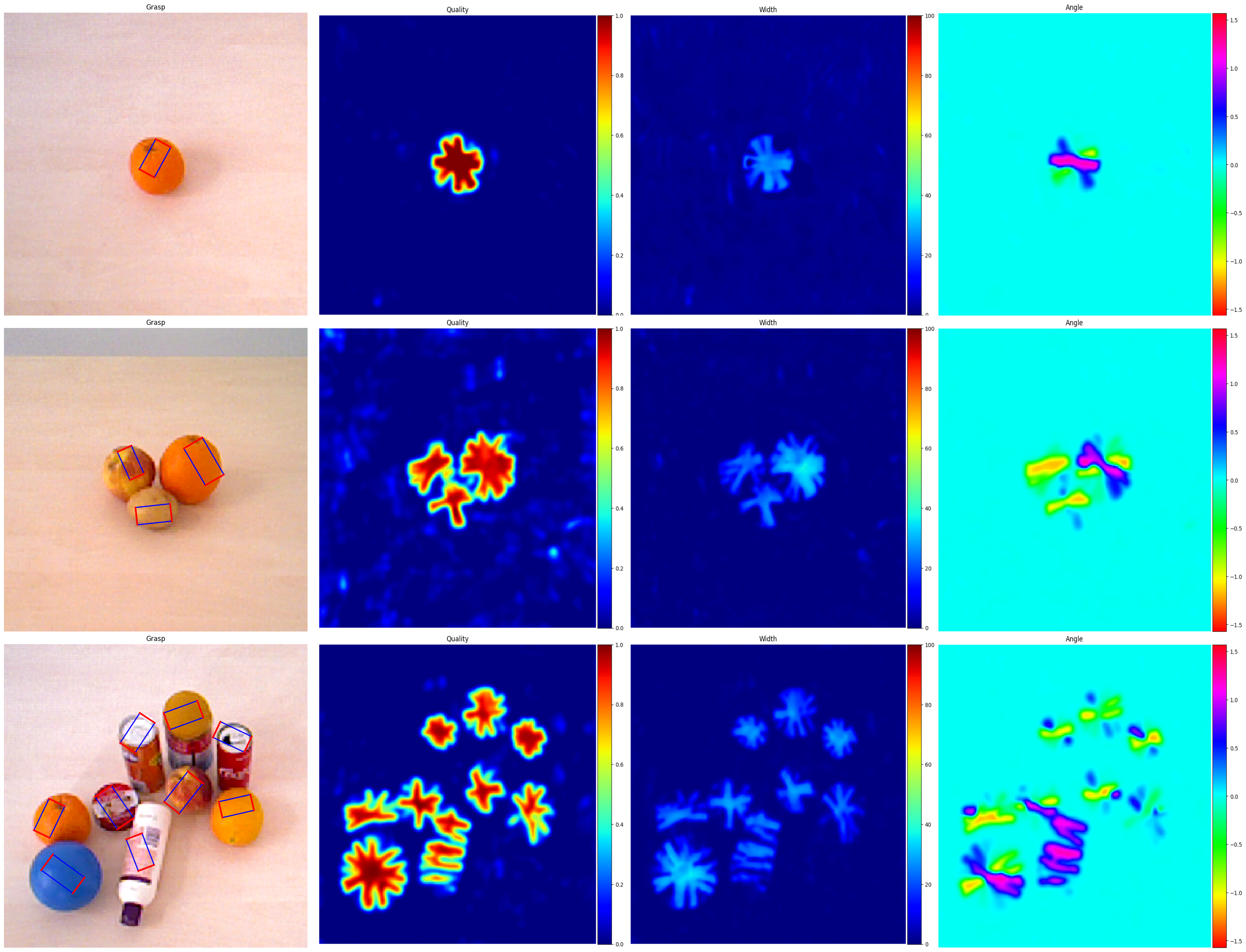}}

\caption{Multiple-object grasp detection using the HMT. From left to right: grasp rectangles on RGB images, and heatmaps for grasp quality, width, and angle.}
\label{multi_object_vis}
\vspace{-5mm}
\end{figure}

We first evaluated our method on the Cornell dataset using 5-fold cross-validation. Table~\ref{tab:cornell} presents the results for both image-wise (IW) and object-wise (OW) splits. Our proposed approach achieved a detection accuracy of 99.55\%, outperforming other state-of-the-art methods in both splits. The superior performance of HMT-Grasp can be attributed to its ability to effectively capture both local and global features, enabling more precise grasp detection even in challenging scenarios. This result illustrates the robustness and effectiveness of our approach across various input data configurations in the Cornell dataset. Additionally, the experimental result also demonstrates that our method achieves good real-time performance on the tested experimental equipment.

\begin{table}[!b]
\centering
\caption{Results of our methods on the Cornell Grasping Dataset in image-wise and object-wise settings.}
\begin{tabular}{l|l|c|c|c}
\hline
\textbf{Method} & \textbf{Input} & \multicolumn{2}{c|}{\textbf{Accuracy (\%)}} & \textbf{Time (ms)} \\ \cline{3-4}
                                        &     & \textbf{IW} & \textbf{OW} &    \\ \hline
GG-CNN~\cite{morrison2020learning}    & D        & 73.0     & 69.0     & 19 \\
GPRN~\cite{karaoguz2019object}       & RGB       & 88.7     & -        & 200\\
HybridGrasp~\cite{guo2017hybrid}     & RGB-D     & 93.2     &89.1     & -  \\
GR-ConvNet~\cite{kumra2020antipodal} & RGB       & 96.6     & 95.5       & 19 \\
GR-ConvNet~\cite{kumra2020antipodal} & RGB-D     & 97.7     & 96.6       & 20 \\
TF-Grasp~\cite{wang2022transformer}   & D        & 95.2     & 94.9      & 41.1 \\
TF-Grasp~\cite{wang2022transformer}  & RGB       & 96.78    & 95.0      & 41.3 \\
TF-Grasp~\cite{wang2022transformer}   & RGB-D    & 97.99   & 96.7      & 41.6 \\
HRG-Net~\cite{zhou2022robotic}   & D    & 99.43   & 96.8      & 52.6 \\
HRG-Net~\cite{zhou2022robotic}   & RGB    & 98.50   & 96.7      & 53.0 \\
HRG-Net~\cite{zhou2022robotic}   & RGB-D    & 99.50   & 97.5      & 53.7 \\
AE-GDN~\cite{qin2023attention}   & RGB   & 97.2   & 96.4   & 24.6 \\
AE-GDN~\cite{qin2023attention}   & RGB-D  & 98.9 & 97.9   & 26.4 \\
DCSFC-Grasp~\cite{zou2023robotic}   & RGB-D  & 99.3 & 98.5   & 22 \\
CGD-CNN~\cite{gu2024cooperative}   & RGB-D  & 97.8 & 96.6   & 19 \\
QQGNN~\cite{fu2024light}   & RGB-D  & 97.7 & 98.9   & 14.7 \\

\hline
\multirow{3}{*}{HMT-Grasp (Ours)}  & RGB    & 99.21 & 99.44 & 23.8 \\                                              
                                 & D & 99.32 & 99.55 & 22.3 \\
                                & RGB-D & 99.55 & 99.55 & 24.8 \\
                                   \hline
\end{tabular}
\label{tab:cornell}
\end{table}

\subsubsection{Grasping Performance on the Jacquard Dataset}

\begin{table}[!t]
\centering
\vspace{2mm}
\caption{The Accuracy on Jacquard Grasping Dataset.}
\begin{tabular}{c|c|c|c}
\hline
\textbf{Reference} & \textbf{Method} & \textbf{Input} & \textbf{Accuracy} (\%) \\
\hline
\multirow{1}{*}{Morrison~\cite{morrison2020learning}} 
                      & GGCNN2 & RGBD & 88.3 \\

\multirow{1}{*}{Kumra~\cite{kumra2020antipodal}} 
                      & GR-ConvNet & RGB-D & 91.0 \\

\multirow{1}{*}{Wang~\cite{wang2022transformer}} 
                      & TF-Grasp & RGB-D & 86.7 \\
\hline
\multirow{3}{*}{Ours} & HMT-Grasp & D & \textbf{93.0} \\
                           & HMT-Grasp & RGB & \textbf{90.4} \\
                           & HMT-Grasp & RGB-D & \textbf{93.3} \\
\hline
\end{tabular}
\vspace{-5mm}
\label{tab:jacquard}
\end{table}

Data splitting for the Jacquard dataset varies across methods. For example, GGCNN2 uses 95\% of the dataset for training, while TF-Grasp and GR-ConvNet use 90\%, with minor differences in test splits. Additionally, distinctions between image-wise and object-wise splits are often unclear. To ensure fair comparisons, we reran all Jacquard experiments following the same protocol as state-of-the-art methods. The training and evaluation protocol is detailed in the \textit{Jacquard Dataset and Training} section. As shown in Table~\ref{tab:jacquard}, our HMT-Grasp approach, using RGB-D input, achieved 93.3\% accuracy, surpassing existing methods. The consistent improvement across different data splits reinforces the effectiveness of our hybrid architecture in capturing the critical features required for successful grasp prediction.

\subsubsection{Grasping Performance on the OCID-Grasp Dataset}

\begin{table}[!b]
\centering
\vspace{-3mm}
\caption{The Accuracy on OCID-Grasp Grasping Dataset.}
\begin{tabular}{c|c|c|c}
\hline
\textbf{Reference} & \textbf{Method} & \textbf{Input} & \textbf{Accuracy} (\%) \\
\hline
\multirow{3}{*}{Morrison~\cite{morrison2020learning}} & GGCNN2 & D & 69.3 \\
                      & GGCNN2~ & RGB & 65.7 \\
                      & GGCNN2 & RGBD & 67.6 \\
\hline
\multirow{3}{*}{Kumra~\cite{kumra2020antipodal}} & GR-ConvNet & D & 70.0 \\
                      & GR-ConvNet & RGB & 68.5 \\
                      & GR-ConvNet & RGB-D & 72.3 \\
\hline
\multirow{3}{*}{Wang~\cite{wang2022transformer}} & TF-Grasp & D & 72.0 \\
                      & TF-Grasp & RGB & 58.9 \\
                      & TF-Grasp & RGB-D & 61.8 \\
\hline
\multirow{3}{*}{Ours} & HMT-Grasp & D & \textbf{73.5} \\
                           & HMT-Grasp & RGB & \textbf{76.5} \\
                           & HMT-Grasp & RGB-D & \textbf{74.1} \\
\hline
\end{tabular}
\label{tab:ocid}
\end{table}

To validate our method in complex environments, we evaluated it on the OCID-Grasp dataset. The results, presented in Table~\ref{tab:ocid}, follow the settings outlined in the \textit{OCID Dataset and Training section}. Our approach, using RGB input, outperformed other methods, demonstrating its superior ability to handle complex grasping scenarios, and showing greater adaptability compared to ViTs and CNNs. As shown in Fig~\ref{multi_object_vis}, the grasping detection results demonstrates our proposed grasp detection method can achieve a good performance across multiple objects scenarios. The superior performance in this cluttered environment underscores the enhanced feature extraction capability of the Mamba blocks, which balance local and global information to enable more effective grasping decisions in difficult scenarios.

\subsection{Ablation Studies }
To further investigate the impact of skip-connections and the Mamba block on grasp pose learning, we trained our network using 5-fold cross-validation on the Cornell dataset with image-wise splitting.
\subsubsection{Ablation study for skip-connections}
We firstly conducted experiments with and without skip-connections using our method across RGB, Depth, and RGB-D inputs. The detailed experimental results are presented in Table~\ref{tab:ablation_skip}.
The study results indicate that incorporating skip connections improves grasp detection performance, particularly for RGB-based inputs, achieving a 2.3$\%$ improvement compared to the absence of skip connections.

\begin{table}[!t]
\centering
\vspace{2mm}
\caption{Ablation experimental results from different skip-connections on the OCID-Grasp dataset.}
\begin{tabular}{|c|c|c|}
\hline
\diagbox{Modality}{Connections} & \makecell{With \\ Skip-connections} & \makecell{Without\\ Skip-connections }\\ 
\hline
Depth(\%) & 73.5 & 73.1  \\
\hline
RGB(\%) & 76.5 & 74.2   \\
\hline
RGB-D(\%) & 74.1 & 73.4  \\
\hline

\end{tabular}
\vspace{-4mm}
\label{tab:ablation_skip}
\end{table}

\subsubsection{Ablation study for different components}

\begin{figure}[!t]
\vspace{1.5mm}
\centerline{\includegraphics[width=\linewidth]{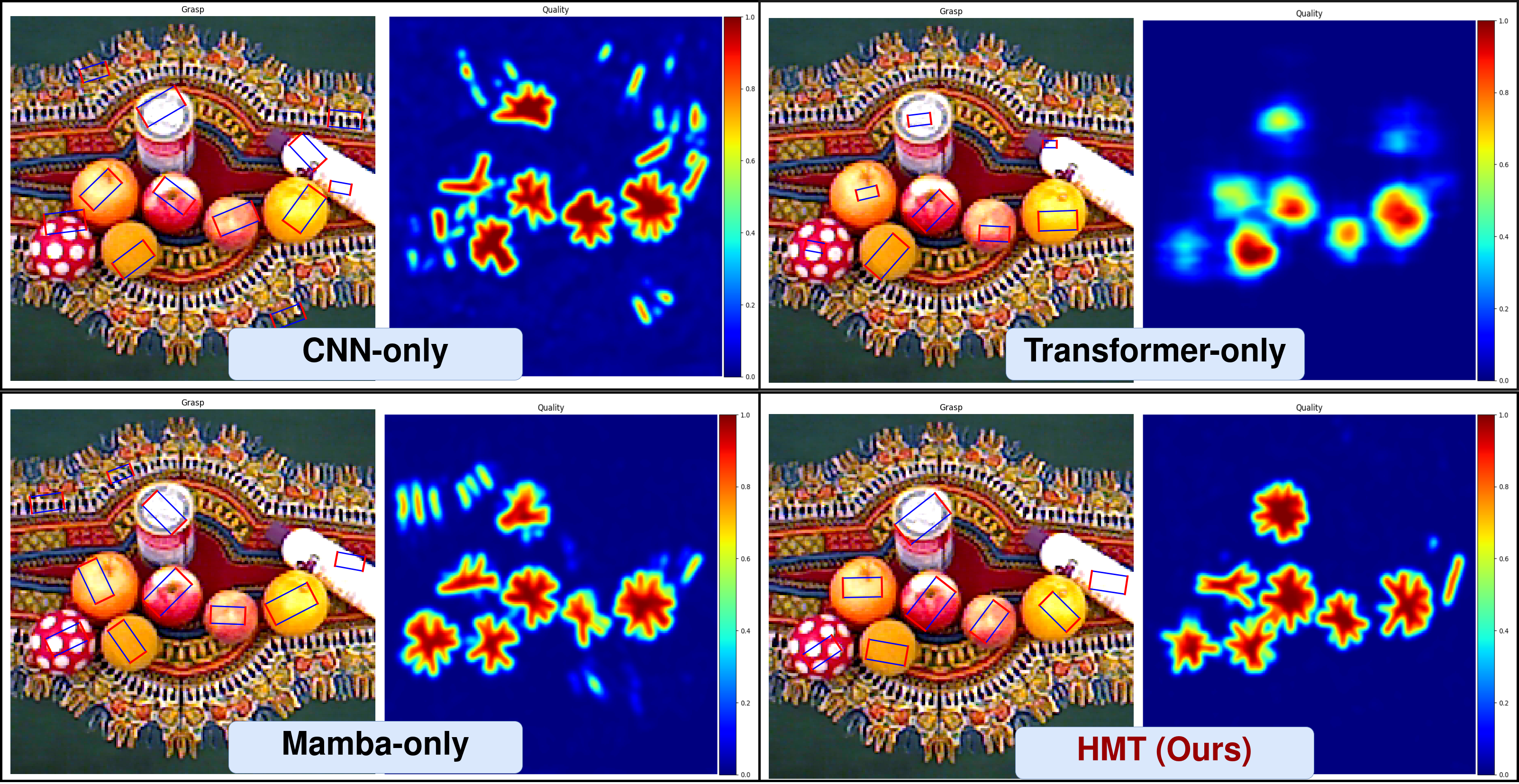}}

\caption{Comparison of grasp detection outcomes and quality across CNN, Transformer, Mamba, and HMT methods in complex multi-object scenes. }
\label{abaltion_detection_vis}
\vspace{-5mm}
\end{figure}

To evaluate the contributions of the CNN, Transformer, and Mamba modules in our proposed method, we conduct an ablation study on the OCID-Grasp dataset. The results are presented in Table~\ref{tab:ablation_components}. As shown in the first row, using CNN-only achieves accuracy of 73.9\%. When replacing the CNN with only the Transformer and Mamba modules, the accuracy drops to 71.2\%, indicating that the convolutional network captures essential local grasping features, such as the textures and edge, as shown in Fig.~\ref{abaltion_detection_vis} that significantly contribute to detection performance. 

When combining the CNN with the Mamba module, the accuracy remains at 73.9\%, showing no substantial improvement compared to using only the CNN or Mamba. As shown in Fig.~\ref{abaltion_detection_vis}, the quality heatmaps of Mamba-only and CNN-only models indicate that both focus more on local details compared to the Transformer-only model. This also suggests that while Mamba effectively enhances sequence modeling, it lacks the capability to fully compensate for the absence of strong global representations, which are crucial in complex, multi-object grasping scenarios where spatial relationships between objects and background diversity play a significant role. In contrast, the combination of the CNN and Transformer results in a slight performance drop compared to the standalone components. This outcome underscores the importance of the Mamba module in balancing local and global representations. The Mamba module refines feature extraction by integrating local information from the CNN and global spatial awareness from the Transformer, ensuring that the model does not overly prioritize one type of representation at the expense of the other. 

Finally, our method, which integrates CNN, Transformer, and Mamba, achieves the highest accuracy of 76.5\%, demonstrating Mamba’s effectiveness in furthering refine local and global features for enhanced grasp detection. These results validate our architecture. CNN primarily captures fine local details, the Transformer focuses on extracting global context, and Mamba effectively fuses and enhances both, ultimately improving overall performance.

\begin{table}[ht]
    \centering
    \caption{Ablation study on the OCID-Grasp dataset illustrating the contribution of the CNN, Transformer, and Mamba modules. }
    \label{tab:ablation_components}
    \begin{tabular}{ccc|c}
        \toprule
        \multicolumn{3}{c|}{\textbf{Ablation Settings}} & \multirow{2}{*}{\textbf{Accuracy(\%)}} \\
        \cmidrule{1-3}
        \textbf{CNN} & \textbf{Transformer} &  \textbf{Mamba} &  \\
        \midrule
        \checkmark &  &  & 73.9  \\
         & \checkmark &  & 74.0  \\
         &  & \checkmark & 74.1 \\
        \checkmark & \checkmark &  & 72.7  \\

         & \checkmark & \checkmark & 71.2 \\
        \checkmark&  & \checkmark & 73.9 \\
        % \checkmark & \checkmark & X & 00\% \\

        \textbf{\checkmark} & \textbf{\checkmark} & \textbf{\checkmark} & \textbf{\hspace{3.5em}76.5} (\textbf{\textcolor{red}{   Ours}}) \\

        \bottomrule
        \vspace{-5mm}
    \end{tabular}
\end{table}

\subsection{Simulation and Real Robot Experiments }

\begin{figure}[!b]
\vspace{-4mm}
\centerline{\includegraphics[width=\linewidth]{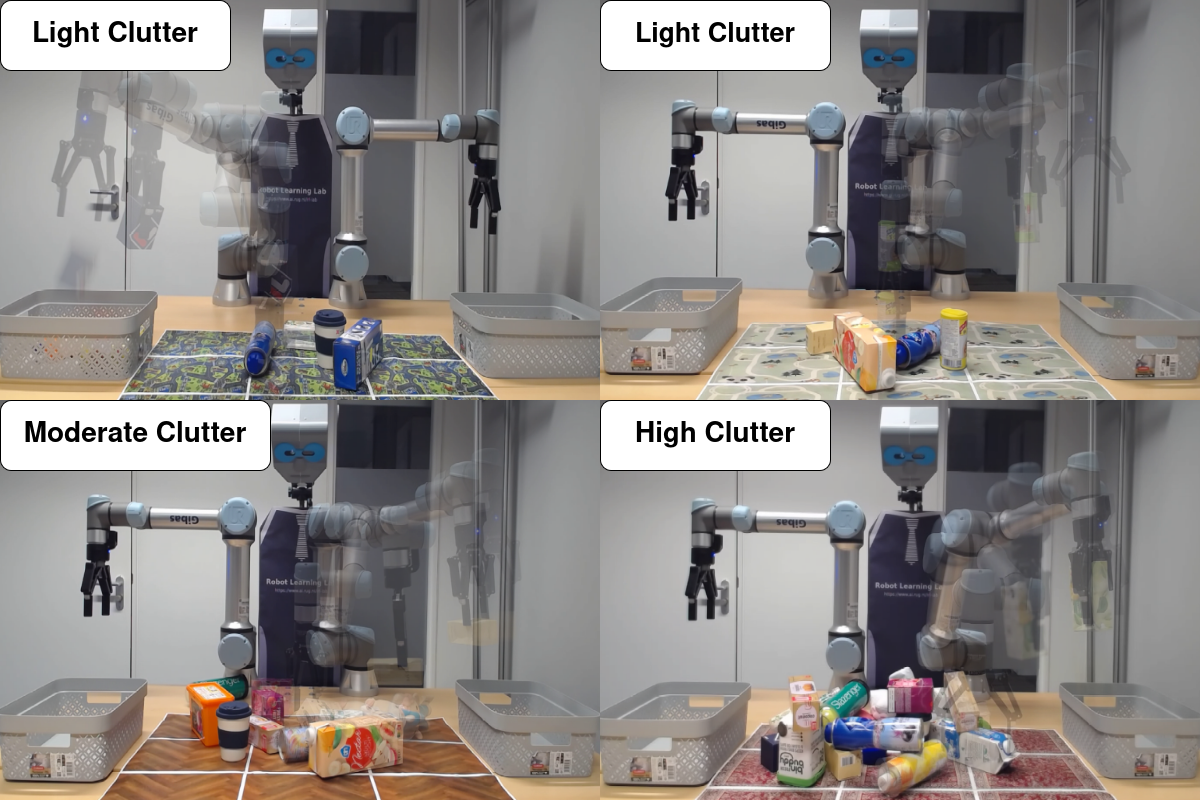}}
\caption{Experimental setups for robotic grasping with varying clutter: Light, Moderate, and High.}
\label{real_exps}
\vspace{-1mm}
\end{figure}

\begin{figure}[!t]
\vspace{1.5mm}
\centerline{\includegraphics[width=\linewidth]{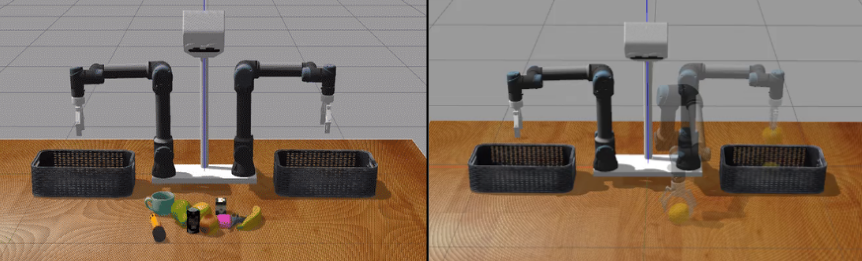}}
\caption{Simulation setups: (\textit{left}) all simulation objects; (\textit{right}) our robot grasps and releases a randomly placed object.}
\label{sim_exps}
\vspace{-0.5mm}
\end{figure}

In this section, we first conducted robot grasping experiments in the Gazebo using models, as shown in Fig.~\ref{sim_exps}. 
First, we performed 20 isolated experiments for each object by randomly spawning it within the robot's workspace. The results, presented in Table~\ref{tab:sim}, demonstrate the effectiveness of our method in grasping various objects. 
The results show that HMT-Grasp generally outperforms other methods across most objects. It achieved 100\% success on the pear, surpassing GR-ConvNet (80\%) and TF-Grasp (85\%). It also performed well on can\_coke (95\%), exceeding all other methods. Although HMT-Grasp generally performed better, there were cases where the improvement was marginal. For example, HMT-Grasp and GR-ConvNet both reached 85\% on the banana, and the performance difference for the peach was only $5\%$. The reasons for failures can be attributed to the complexity of object shapes, material and collision mesh properties. For example, objects like the mug had lower success rates for both methods. The challenge with grasping this object lies in the difficulty of detecting stable grasp points, as small surface areas reduce friction and stability.

We conducted real-robot experiments in lightly, moderately, and heavily cluttered environments with complex backgrounds, as illustrated in Fig.~\ref{real_exps}. In the lightly cluttered setting, we tested twenty groups of five randomized objects per method to ensure diverse grasping scenarios. In the moderately cluttered setting, we conducted ten groups of ten randomized objects per method to evaluate grasping performance across various backgrounds. In the highly cluttered setting, we ran five experimental groups, each aiming to include twenty distinct objects to assess robustness under dense clutter. We compute the success rate as the ratio of successful grasps to the total number of grasp attempts. The results are presented in Table~\ref{tab:real}. Overall, our approach consistently showed better performance compared to the baseline methods. However, accuracy decreased from lightly to heavily cluttered scenarios for two primary reasons: (1) insufficient friction between the gripper jaws and certain smooth-surfaced objects, resulting in grasp failures; and (2) the inaccurate grasp prediction led to contacting the gripper with the neighboring objects, resulting in a grasp failure.

\begin{table}[ht]
    \centering
\caption{Simulation results: 20 simulations were performed for each object to assess the grasp success rate ($\%$)}
    \begin{tabular}{|c|c|c|c|c|c|}
        \hline
        Objects & Pear & lemon & sponge & can\_coke & Peach \\
        \hline
        GR-ConvNet~\cite{kumra2020antipodal}  & 80 & 75 & 85 & 80 & 75 \\
        TF-Grasp~\cite{wang2022transformer}  & 85 & 70 & 70 & 85 & 70 \\
        \textbf{HMT-Grasp}  & 100 & 95 & 75 & 95 & 80 \\
        \hline
        Objects & mug & scissor & tea\_box & banana & flash \\
        \hline
        GR-ConvNet~\cite{kumra2020antipodal}  & 80 & 70 & 70 & 85 & 75 \\
        TF-Grasp~\cite{wang2022transformer}  & 75 & 85 & 80 & 75 & 80 \\
        \textbf{HMT-Grasp}  & 70 & 95 & 80 & 85 & 90 \\
        \hline
    \end{tabular}
    \label{tab:sim}
\end{table}

\begin{table}[ht]
    \centering
    \vspace{-2mm}
    \caption{Real-world robot grasp success rates in lightly , moderately, and highly cluttered scenarios.}
    \begin{tabular}{|c|c|c|c|}
        \hline
        \multirow{3}{*}{\textbf{Models}} & \multicolumn{3}{c|}{\textbf{Success Rate ($\%$)}} \\ \cline{2-4}
        & Light Clutter & Moderate Clutter & High Clutter\\  
        & (5 Objects) & (10 Objects)& (20 Objects) \\ \hline
        
        GR-ConvNet~\cite{kumra2020antipodal} & 80 & 74 & 72 
        \\ \hline
        TF-Grasp~\cite{wang2022transformer} & 82 & 75 & 70 
        \\ \hline

        \textbf{HMT-Grasp} & 88 & 82 & 79 \\ \hline
    \end{tabular}
    \vspace{1mm}
    \label{tab:real}
\end{table}

\section{Conclusion}
This paper presents a hybrid Mamba-Transformer model for robotic grasping in complex scenarios. By combining Mamba, Transformers, and CNNs, our method improves the performance of ViTs and CNNs in visual grasp detection. We conducted extensive evaluations on the Cornell, Jacquard, and OCID-Grasp datasets, where our model consistently outperformed state-of-the-art methods, achieving an accuracy of 99.5\% on the Cornell dataset. These results demonstrate the superior adaptability and precision of our approach. Additionally, real-world robotic experiments further validated the method's effectiveness in complex environments, highlighting its potential for practical applications. Future work could focus on incorporating real-time feedback mechanisms to adjust grasp plans dynamically during execution.

\bibliographystyle{IEEEtran}
\small\bibliography{reference}

\end{document}